\documentclass{article}
\usepackage{ijcai25}

\usepackage{times}
\usepackage{soul}
\usepackage{url}
\usepackage[hidelinks]{hyperref}
\usepackage[utf8]{inputenc}
\usepackage[small]{caption}
\usepackage{graphicx}
\usepackage{amsmath}
\usepackage{amsthm}
\usepackage{booktabs}
\usepackage{algorithm}
\usepackage{algpseudocode}
\usepackage[switch]{lineno}
\usepackage{tcolorbox}
\usepackage{multirow}
\usepackage{mathrsfs}
\usepackage{subfig}
\usepackage{hyperref} 
\usepackage{amssymb}
\usepackage{bm}
\usepackage{newfloat}
\usepackage{listings}
\usepackage{algpseudocode}
\usepackage[top=2cm, bottom=2cm, left=2cm, right=2cm]{geometry}
\floatname{algorithm}{Algorithm}
\newtheorem{Assumption}{Assumption}
\newtheorem{Definition}{Definition}
\newtheorem{Theorem}{Theorem}
\newtheorem{Lemma}{Lemma}

\allowdisplaybreaks[4]
\newfloat{listing}{tb}{lst}{}
\floatname{listing}{Listing}
\title{Testing for Causal Fairness}

\author{
Jiarun Fu$^1$
\and
Lizhong Ding$^1$\and
Pengqi Li$^{1}$\and
Qiuning Wei$^1$\and
Yurong Cheng$^1$\and
Xu Chen $^2$
\affiliations
$^1$Beijing Institute of Technology\\
$^2$Renmin University of China\\
\emails
3120235197@bit.edu.cn
}

\begin{document}

\maketitle

\begin{abstract}
Causality is widely used in fairness analysis to prevent discrimination on sensitive attributes, such as genders in career recruitment and races in crime prediction. However, the current data-based Potential Outcomes Framework (POF) often leads to untrustworthy fairness analysis results when handling high-dimensional data. To address this, we introduce a distribution-based POF that transform fairness analysis into Distributional Closeness Testing (DCT) by intervening on sensitive attributes. We define counterfactual closeness fairness as the null hypothesis of DCT, where a sensitive attribute is considered fair if its factual and counterfactual potential outcome distributions are sufficiently close. We introduce the Norm-Adaptive Maximum Mean Discrepancy Treatment Effect (N-TE) as a statistic for measuring distributional closeness and apply DCT using the empirical estimator of NTE, referred to  Counterfactual Fairness-CLOseness Testing ($\textrm{CF-CLOT}$). To ensure the trustworthiness of testing results,  we establish the testing consistency of N-TE through rigorous theoretical analysis. $\textrm{CF-CLOT}$ demonstrates sensitivity in fairness analysis through the flexibility of the closeness parameter $\epsilon$. Unfair sensitive attributes have been successfully tested by $\textrm{CF-CLOT}$ in extensive experiments across various real-world scenarios, which validate the consistency of the testing.

\end{abstract}

\section{Introduction}

With the rapid advancement of machine learning technologies, AI-based decision-making models have become instrumental in addressing some of society's most pressing challenges, including healthcare, law enforcement, education, and finance \cite{2018Forecasting,2022Forecasting,xue2023prompt,melnychuk2024partial}. However, concerns about the fairness of these models have grown significantly from both social and ethical perspectives. Notable examples include the COMPAS system, which disproportionately discriminates against African Americans in U.S. recidivism prediction \cite{dressel2018accuracy}, and corporate decision-making systems that exhibit bias against female employees in career recruitment \cite{salimi2019interventional,glymour2019review}. The existence of discrimination in large language models \cite{balestri2024examiningmultimodalgendercontent} further highlights the urgency of fairness. Consequently, it is important to measure and alleviate the unfairness of models. The research community has proposed a variety of methods to meausre the fairness of deep models \cite{taskesen2021statistical,mehrabi2021survey,jiang2024chasing}. These methods are often considered to fall within the broader domain of trustworthy machine learning \cite{zhu2024survey}, is urgently needed for us due to its importance in ensuring that automated decisions align with ethical standards \cite{wen2023survey,xie2024local}.

Current fairness measures can generally be classified into two categories: statistical-based measures and causal-based ones \cite{mehrabi2021survey}. Statistical fairness measures is widely used due to its clear standards for quantifying fairness \cite{chen2019fairness,taskesen2021statistical,jiang2024chasing}. But they are often prone to errors in complex data scenarios, a phenomenon known as ``statistical illusion'' \cite{yao2021survey,kaddour2022causal,raghavan2024limitations}, which can lead to unintended consequences and bias amplification \cite{plecko2022causalfairnessanalysis}. In contrast, causal-based measures focus on estimating the causal effects of sensitive attributes to evaluate fairness and avoid statistical bias \cite{zuo2023counterfactualfairnesspartiallyknown,li2024debiased}. In Potential Outcome Framework (POF), it is widely believed that, for fair models, there should be no causal relationship between sensitive attributes and outcomes. For instance, there should be no causal relationship between races and crime rates, or between genders and employment rates. Currently, fairness analysis primarily employs data-based POF, which measures causal effects by intervening on sensitive attributes and calculating the expected difference between factual and counterfactual potential outcomes \cite{rubin1980randomization,pearl2014interpretation,plecko2022causalfairnessanalysis}. But with growing societal concerns about data privacy and security, \cite{huang2022large,wei2024jailbroken}, obtaining data with all attributes is nearly impossible. This results in sparse samples, where many attribute combinations may lack sufficient data support \cite{wu2023stable}. As a consequence, data-based POF struggles to clearly analyze interaction effects in high-dimensional data \cite{zhang2024stableheterogeneoustreatmenteffect} (The outcomes are considered to require validation across all continuous bounded functions, see Appendix~\ref{into}.), leading to untrustworthy estimates of potential outcome samples and, in turn, compromising the accuracy of fairness analysis.

Fortunately, kernel based distrubition testing can effectively reflect differences between samples\cite{liu2016kernelized}, as demonstrated in famous two-sample testing \cite{gretton2012kernel}. Reproducing kernel Hilbert space (RKHS) uses characteristic kernels to map data into a high-dimensional space, transforming potentially complex distribution differences into a linear problem. It then quantifies the gap between two distributions by calculating the difference in their mapped mean values in the high-dimensional space \cite{hofmann2008kernel,cho2009kernel,gretton2012kernel}.

To address the limitation of data-based POF fairness analysis, we introduce distribution-based POF, transform fairness analysis into Distributional Closeness Testing (DCT). To equate closeness with fairness, we define counterfactual closeness fairness where the factual potential and counterfactual potential outcome distributions are close within a closeness parameter $\epsilon$ by intervening on sensitive attributes. We apply Norm-adaptive maximum mean discrepancy (NAMMD) \cite{anonymous2024a} to propose our statistic, the NAMMD treatment effect (N-TE), achieving precise measurement of the closeness between distributions based on RKHS. By defining the sufficient closeness of factual and counterfactual potential outcome distributions under the empirical estimator of N-TE as the null hypothesis for fairness of sensitive attributes, we propose our method, Counterfactual Fairness Closeness Testing ($\textrm{CF-CLOT}$). Current kernel-based distribution tests \cite{gretton2012kernel,liu2016kernelized}, only recognize fairness when the distributions are completely identical. It does not meet the sensitivity requirements for fairness analysis. For example, if a large language model (LLM) rates sensitive photos of men at 73\% and women at 73.5\%, the two-sample test would label it as unfair, even though there is no significant gender discrimination in general cognition. Since N-TE normalizes the treatment effects to lie within the [0, 1] range, we can interpret its results as the degree of distributional differences corresponding to the RKHS. Therefore, 
$\textrm{CF-CLOT}$ exhibits sensitivity in fairness analysis due to the flexible and reasonable $\epsilon$. We establish the testing consistency of N-TE through rigorous theoretical analysis to ensure unfair sensitive attributes can be successfully tested by $\textrm{CF-CLOT}$. 

Our contributions can be summarized as follows:
\begin{enumerate}
    \item For the first time, we introduce distribution-based POF to transform fairness analysis into DCT. We define counterfactual closeness fairness as the null hypothesis to establish a link between closeness and fairness.
    \item We define N-TE as a statistic that measures the treatment effect from the perspective of distributional closeness. The testing consistency of N-TE has been thoroughly analyzed to ensure unfair sensitive attributes can be trustworthily tested.
    \item Our method $\textrm{CF-CLOT}$ demonstrates sensitivity to fairness analysis through the flexibility of fairness confidence $\epsilon$ and we validate it through real-world experiments.
\end{enumerate}

\section{Preliminaries}

\subsection{Potential outcomes causal framework}

As a starting point, the Potential Outcomes Framework (POF) \cite{rubin1980randomization,pearl2009causality} focuses on a treatment with two levels. We use $T \in {0,1}$ to represent the treatment levels, where $T = 1$ indicates treatment and $T = 0$ indicates no treatment. Let $Y$ represents the outcome of interest, and $\hat{Y}$ denotes the predicted value of $Y$. $\hat{Y}$ has a factual version, $(\hat{Y} | T = 0)$ (which we will refer to it as $\hat{Y}$ for simplicity), and an counterfactual version, $(\hat{Y} | T = 1)$, corresponding to the hypothetical treatment interventions $T = 0$ and $T = 1$, respectively. The fundamental problem of causal inference is defined as evaluating the difference between $(\hat{Y} | T = 1)$ and $\hat{Y}$. This difference is known as the treatment effect, and if the treatment effect exists, it indicates a causal relationship between the treatment and the outcome.

\subsection{Distributional closeness testing}
DCT accesses whether two unknown discrete distributions are close from each other based on the selected discrepancy measure \cite{li1996nonparametric,canonne2020survey}. Over domain $\mathcal Z=\{\boldsymbol z_1,\boldsymbol{z}_2,\ldots,\boldsymbol{z_n}\}\subseteq\mathbb{R}^d$, let $\mathbb{P}_n=\{p_1,p_2,...,p_n\}$ and $\mathbb{Q}_n=\{q_1,q_2,...,q_n\}$ be two discrete distributions, such that $\sum_{i=1}^np_i=1$ and $\sum_{i=1}^nq_i=1$, respectively. Let $\kappa:\mathcal{Z} \times \mathcal{Z} \rightarrow \mathbb{R}$ be the PDS kernel with respective to corresponding reproducing kernel Hilbert space $(\mathcal{H_{\kappa}},\langle\cdot,\cdot\rangle_{\mathcal{H_{\kappa}}})$, where $\kappa(\cdot,z_i)\in \mathcal{H_{\kappa}}$ and $\langle \kappa(\cdot,z_i),\kappa(\cdot,z_j)\rangle=\kappa(z_i,z_j)$. We first define the kernel mean embedding of $\mathbb P_n$ and $\mathbb Q_n$ as:
\begin{equation}
\label{7}
    \mu({\mathbb{P}_n})=E_{z\sim\mathbb P_n}\left[\kappa(\cdot,z)\right],\ \mu({\mathbb Q}_n)=E_{z\sim\mathbb Q_n}\left[\kappa(\cdot,z)\right].
\end{equation}
We assume that there exists $K>0$, such that $0	
\leq \kappa(z_i,z_j) \leq K$ for any $z_i,z_j\in\mathcal Z$. We use the inner product induced norm $\|\cdot\|_{\mathcal H_\kappa}$ to measure the discrepancy of $\mathbb{P}_n$ and $\mathbb{Q}_n$:
  \begin{equation}
 \begin{aligned}
\text{D}_\kappa(\mathbb{P}_n , \mathbb{Q}_n)=\|\mu({\mathbb{P}_n})-\mu({ \mathbb{Q}_n})\|_{\mathcal H_\kappa}.
 \end{aligned}
 \end{equation}
Taking $\text{D}_\kappa$ as the measure of distribution closeness, the goal of Distribution Closeness Testing (DCT) is to test between the null and alternative hypotheses as follows:
\begin{equation}
H_{0} : \text{D}_\kappa(\mathbb{P}_n, \mathbb{Q}_n)\leq \epsilon, H_{1} : \text{D}_\kappa(\mathbb{P}_n , \mathbb{Q}_n)\textgreater \epsilon,
\end{equation}
where $\epsilon$ denotes the predetermined closeness parameter.

\section{Problem Formulations and Assumptions}

\subsection{Problem formulations}

As widely recognized in fairness analysis, let $A$ represent the set of all attributes, with $A_s$ and $A_c$ denoting the sensitive and non-sensitive (observable) attributes, respectively. The core idea of causal fairness analysis is to test whether $A_s$ has a causal effect on the outcome of interest, $Y$. If such a causal effect exists, it indicates that the model is unfair; otherwise, the model is considered fair. After selecting a sensitive attribute $a_{\mathrm{testing}} \in A_s$ for testing (for simplicity, we will denote $a_{\mathrm{testing}}$ as $a_t$ hereafter), we perform interventions using the do-operator, $do(\cdot)$ (the intervention process generally involves either removing sensitive attributes or converting them into noise that does not influence the outcome). Let $do(a_t \rightarrow a_t')$ represent the intervention under treatment $T=1$, where $a_t'$ denotes the counterfactual version of the sensitive attribute. The corresponding counterfactual outcome is denoted by $\hat{Y}_{a_t'}$. Now, assume that the random variable $\hat{Y}$ follows a discrete distribution $\mathbb{P}_n$, and the corresponding counterfactual random variable $\hat{Y}{a_t'}$ follows the distribution $\mathbb{P}_{a_t',n}$, then:

\begin{Definition}[Distribution-based POF] 
\label{DPOF}
Given the deep model as $f_Y$ with inputs from both $A_c$ and $A_s$,  we can express the distribution-based Potential Outcomes Framework (POF) as
\begin{equation}
\begin{aligned}
&\hat{Y}=f_Y(A_s,A_c)\sim\mathbb P_n, \\
&\hat{Y}_{a_t'}=f_Y\big((A_s \backslash \{a_t\}),do(a_t\rightarrow a_t'),A_c\big)\sim\mathbb P_{a_{t}',n}.
\end{aligned}
\end{equation}
\end{Definition}

We aim to complete the causal fairness analysis by testing the causal relationship between sensitive attributes $A_s$ and the outcome of interest $Y$ based on the distribution-based POF. Specifically, we test the treatment effect of $do(a_t \rightarrow a_t')$. The treatment effect is typically calculated as the difference between the realized $\hat{Y}$ and $\hat{Y}_{a_t'}$. To ensure the trustworthiness of our testing, we measure the treatment effect by assessing the closeness between the distributions $\mathbb{P}_n$ and $\mathbb{P}_{a_t',n}$, which is defined as follows.

\begin{Definition}[Counterfactual distribution closeness] 
\label{IDC}
We use discrepancy to measure the closeness between $\mathbb P_n$ and $\mathbb P_{a'_t,n}$:

\begin{equation}
\begin{aligned}
\mathrm{D}_\kappa(\mathbb P_n, \mathbb P_{a'_t,n})=\|\mu({\mathbb P_n})-\mu({\mathbb P_{a_t',n}})\|_{\mathcal H_\kappa}.
\end{aligned}
\end{equation}
\end{Definition}

If we follow the conventional definition of causality, $\text{D}_\kappa(\mathbb P_n, \mathbb P_{a'_t,n})>0$ means $a_t'$ has a causal relation to $Y$, and $f_Y$ will be regarded unfair to $a_t'$. But such definition is too strict in causal fairness analysis. For example, in a recruitment scenario, due to differences in job attributes, a 5\% difference in recruitment pass rate between genders is acceptable \cite{glymour2019review}. However, in the pass rate generated by a large language model, such a difference is already considered significant \cite{balestri2024examiningmultimodalgendercontent}. Therefore, based on Definition~\ref{IDC} and DCT, we define counterfactual closeness fairness as follows:

\begin{Definition}[Counterfactual closeness fairness]
\label{icf}
A deep model $f_Y(\cdot)$ is counterfactual fair \cite{kusner2017counterfactual,salimi2019interventional} if Eq~\eqref{eq:ICF} holds for all sensitive attributes $a_t\in A_s$:
 \begin{equation}
 \label{eq:ICF}
 \begin{aligned}
  &\Pr\left(\hat{Y}=y\mid A_s,A_c \right)\\
  =&\Pr\left(\hat{Y}_{a_t'}=y\mid (A_s \backslash \{a_t\}),do(a_t\rightarrow a_t'),A_c \right).
 \end{aligned}
 \end{equation}
For $\forall\epsilon\in[0,1)$, the deep model $f_Y(\cdot)$ is of $\epsilon$-counterfactual closeness fairness if $H_{0}$ holds for any $a_t\in A_s$:
 \begin{equation}
H_{0} : \mathrm{D}_\kappa(\mathbb P_n, \mathbb P_{a'_t,n})\leq \epsilon,\quad H_{1} : \mathrm{D}_\kappa(\mathbb P_n, \mathbb P_{a'_t,n})\textgreater \epsilon.
\end{equation}
\end{Definition}

We carefully note that the two $\Pr$ in Eq~\eqref{eq:ICF} differ because the first represents the observational $\mathbb P_n$ while the second represents the counterfactual $\mathbb P_{a_t',n}$. We treat $\epsilon$ as the fairness confidence in Definition~\ref{icf}, which means if the discrepancy between $\mathbb P_n$ and $ \mathbb P_{a_t',n}$ is lower than $\epsilon$, $a_t'$ won't have a causal relation to $Y$, and $f_Y$ will be regarded as fair to $a_t'$. 
\subsection{Assumption}

In this section, according to widely acceptances in the field of causality \cite{xie2020generalizedindependentnoisecondition,wu2023stable,litwo,zeng2023surveycausalreinforcementlearning}, we make three assumptions for our definitions.

\begin{Assumption}[Stable Unit Treatment Value (SUTVA)]
$\mathbb P_n$ and $\mathbb P_{a_t',n}$ with a deep model $f_Y(\cdot)$ are assumed to be independent of interventions $do(a_t\rightarrow a_t')$.
\end{Assumption}

SUTVA ensures Definition ~\ref{IDC} can reflect the correct causal relation between $a_t$ and $Y$ in $f_Y(\cdot)$ since  $\mathbb P_n$ and $\mathbb P_{a_t',n}$ won't change with $do(a_t\rightarrow a_t')$ \cite{qi2023proximal,wu2023stable,zhang2024causaldistillationalleviatingperformance}.

\begin{Assumption}[Unconfoundedness]
The probability distribution $do(a_t\rightarrow a_t')$ is independent of the $\mathbb P_n$ and $\mathbb P_{a_t',n}$ when given $A_c$, and the deep model $f_Y(\cdot)$.
\end{Assumption}

Our second assumption extends the unconfoundedness assumption to Definition ~\ref{icf}, assuming that intervention is not determined by the attributes being tested and all intervention types are equally possible.\cite{qi2023proximal,zuo2024interventional}.

\begin{Assumption}[Overlap]
Every sensitive attribute should have a nonzero probability to receive either treatment status. Formally, $0 < 
\Pr(do(a_t\rightarrow a_t')|f_Y(\cdot),A_c) < 1$.
\end{Assumption}

The overlap assumption is fundamental in the field of causal inference, which ensures the validity and feasibility of do-operator $do(\cdot)$ to $a_t$ in Definition~\ref{icf} \cite{litwo,zeng2023surveycausalreinforcementlearning}.

\section{The Proposed Framework}
In this section, we will provide a detailed explanation of how our method $\textrm{CF-CLOT}$ transform fairness analysis into DCT based on distribution-based POF. We introduce the statistic N-TE to measure the distributional closeness. Then, based on the asymptotic Gaussian distribution, we conduct a rigorous theoretical analysis to establish the testing consistency of the N-TE. Then we propose our method by defining the sufficient closeness of factual and counterfactual potential outcome distributions under the empirical estimator of N-TE as the null hypothesis. In the second section, we will further discuss the sensitivity of $\textrm{CF-CLOT}$ in fairness analysis.

\subsection{Counterfactual fairness-closeness testing}

Obviously, as the deep model $f_Y(\cdot)$ has been determined, whatever $a_t$ is, random variables $\hat{Y}$ and $\hat{Y}_{a_t'}$ are defined on the same instance space $\mathcal{X}\subseteq \mathbb{R}^d$. Then we independently sample $\hat{y},\hat{\boldsymbol{y}} \sim \mathbb P_n $ and $\hat{y}_{a_t'},\hat{\boldsymbol y}_{a_t'} \sim \mathbb P_{a_t',n} $, where $\boldsymbol{\hat{y}}$ (resp. $\boldsymbol{\hat{y}}_{a_t'}$) is an i.i.d. copy of $\hat{y}\sim \mathbb P_n$ (resp. $\boldsymbol{\hat{y}}\sim \mathbb P_{a_t',n}$).

Current kernel based distribution testing method has a limitation that the same discrepancy value may reflect different levels of closeness. To trustworthily test unfair sensitive attributes by DCT, we apply NAMMD \cite{anonymous2024a} to construst NAMMD treatment effect (N-TE) as a statistic:

\begin{Definition}[NAMMD treatment effect (N-TE)] 
\label{NTE}
Given a deep model $f_Y(\cdot)$ and sensitive attribute $a_t\in A_s$, suppose $\hat{Y}$ represents factual potential outcomes and $\hat{Y}_{a_t'}$ represents the counterfactual one. The $\mathrm{N}$-$\mathrm{TE}$ of $do(a_t\rightarrow a_t')$ is
\begin{equation}
\begin{aligned}
&\mathrm{N}\text{-}\mathrm{TE}(f_Y(\cdot),do(a_t\rightarrow a_t'))\\
=&\frac{E_{\hat Y,\hat Y_{a_t'}}[\kappa(\hat{y},\boldsymbol{\hat{y}})+\kappa(\hat{y}_{a_t'},\boldsymbol{\hat{y}}_{a_t'})-2\kappa(\hat{y},\hat{y}_{a_t'})]}{4K-E_{\hat Y,\hat Y_{a_t'}}[\kappa(\hat{y},\boldsymbol{\hat{y}})+\kappa(\hat{y}_{a_t'},\boldsymbol{\hat{y}}_{a_t'})]},
\end{aligned}
\end{equation}
\end{Definition}
\noindent and it is also clear that $\text{N-TE} \in [0, 1]$, so N-TE can be regarded as the degree of the distributional closeness. Based on Definition ~\ref{icf}, the value of N-TE approaches 0 when the $f_Y(\cdot)$ is fair to $a_t$, which makes closeness identical to fairness.

In real-world scenarios, the potential distribution of outcomes is often unknown. A more general perspective is that the potential outcome distributions, $\mathbb{P}_n$ and $\mathbb{P}_{a_t',n}$, represent two groups of independent and identically distributed (i.i.d.) samples. Since $do(a_t \rightarrow a_t')$ only intervenes on the selected sensitive attribute $a_t$, the number of samples in both sets should be equal. This assumption aligns with those made in several similar studies \cite{gretton2012kernel,liu2016kernelized,anonymous2024a}. We express the discrete form of $\hat{Y}$ and $\hat{Y}_{a_t'}$ as follows:

\begin{equation}
\begin{aligned}
\bar Y=\{\hat{y}_i\}_{i=1}^m \sim \mathbb P_n^m ,\bar Y_{a_t'}=\{ {\hat{y}_{a_t',i}} \} _{i=1}^m \sim \mathbb P_{a_t',n}^m.
\end{aligned}
\end{equation}

We further introduce the empirical estimator of N-TE as follows:

\begin{equation}
    \begin{aligned}
&\overline{\text{N-TE}}(f_Y(\cdot),do(a_t \rightarrow a_t'))\\
=&\frac{\sum_{i\neq j} H_{i,j}}
{\sum_{i\neq j}[4K-\kappa(\hat{y}_i,\hat{y}_j) - \kappa({\hat{y}}_{a_t',i},\hat{y}_{a_t',j})]},\\
& 
    \end{aligned}
\end{equation}
for brevity, we use the notation $H_{i, j}=\kappa(\hat{y}_i,\hat{y}_j)+\kappa({\hat{y}}_{a_t',i},{\hat{y}}_{a_t',j})-\kappa(\hat{y}_i,\hat{y}_{a_t',j})-\kappa(\hat{y}_{a_t',i},\hat{y}_j)$.

As determining the type of unknown distribution is difficult, we apply the bootstrap method to the empirical estimator of the N-TE to address this issue \cite{chwialkowski2014wild} by simulating limiting distributions. This approach allows the model to account for data variability by generating multiple resampled datasets, leading to more robust and unbiased estimations. Specifically, we repeatedly draw multinomial random weights $W=(w_1,...,w_m)\sim \mathrm{Mult}(m;\frac{1}{m},...,\frac{1}{m})$ and calculate bootstrap sample as :
\begin{equation}
\label{26}
\begin{aligned}
&\overline{\mathrm{N}\text{-}\mathrm{TE}}_{\phi}(f_Y(\cdot),do(a_t \rightarrow a_t'))=\\
&\frac{\sum_{i\neq j} \phi_{i,j} H_{i,j}}
{\sum_{i\neq j}[4K-\kappa(\hat{y}_i,\hat{y}_j) - \kappa({\hat{y}}_{a_t',i},\hat{y}_{a_t',j})]},\\
& 
    \end{aligned}
\end{equation}

\noindent where $\phi_{ij}=\left( w_i-\frac{1}{m} \right) \left(w_j-\frac{1}{m} \right)$ captures the combined deviation of the two samples from the expected uniform weights. We take $\overline{\text{N-TE}}_{\phi}$ as the measure of our method and define counterfactual fairness-closeness hypothesis in following Definition~\ref{def:CCFH} as the basic for testing.

\begin{Definition}[Counterfactual fairness-closeness
hypothesis]
\label{def:CCFH}
We say $f_Y(\cdot)$ is fair to sensitive attribute $a_t$ if $H_{0}$ holds for any $a_t\in A_s$ with significance level $\alpha$:
\begin{equation}
\begin{aligned}
&H_{0} : {\mathrm{N}\text{-}\mathrm{TE}}\left(f_Y(\cdot),do(a_t \rightarrow a_t')\right)\leq \epsilon,\\
&H_{1} : {\mathrm{N}\text{-}\mathrm{TE}}(f_Y(\cdot),do(a_t \rightarrow a_t')) \textgreater \epsilon.
\end{aligned}
\end{equation}
\end{Definition}

We show that the empirical estimator of our N-TE, i.e. $\text{N-TE}=\epsilon$ has an asymptotic gaussian distribution (The proof will be showed in Appendix~\ref{subsec:A5}).
\begin{Theorem}[$\text{N-TE}$ asymptotic gaussian distribution.]
\label{thm:asy}
If ${\mathrm{N}\text{-}\mathrm{TE}}\left(f_Y(\cdot),do(a_t \rightarrow a_t')\right)=\epsilon$ with $\epsilon\in(0,1]$, we have 
\begin{equation*}
        \sqrt{m}(\overline{\mathrm{N}\text{-}\mathrm{TE}}\left(f_Y(\cdot),do(a_t=a_t')\right)-\epsilon) \xrightarrow{d} \mathcal{N}\left(0, \sigma_{\hat Y,\hat Y_{a_t'}}^2\right),
    \end{equation*}
    where 
    \[
    \begin{aligned}
        \sigma_{\hat Y,\hat Y_{a_t'}}^2=\frac{\sqrt{4 E\left[H_{1,2} H_{1,3}\right]-4\left(E\left[H_{1,2}\right]\right)^{2}} }{\left(4 K-\left\|\boldsymbol{\mu}_{\hat Y}\right\|_{\mathcal{H}_{\kappa}}^{2}-\left\|\boldsymbol{\mu}_{\hat Y_{a_t'}}\right\|_{\mathcal{H}_{\kappa}}^{2}\right)}
    \end{aligned}\]
    and $H_{i, j}=\kappa(\hat{y}_i,\hat{y}_j)+\kappa({\hat{y}}_{a_t',i},{\hat{y}}_{a_t',j})-\kappa(\hat{y}_i,\hat{y}_{a_t',j})-\kappa(\hat{y}_{a_t',i},\hat{y}_j) $ as defined earlier.
\end{Theorem}
Hence, we use the $(1-\alpha)$-quantile of asymptotic
distribution as the testing threshold, denoted by $\tau_\alpha$. Here, the term $\sigma_{\hat Y,\hat Y_{a_t'}}^2$ is unknown in practice and we use the empirical estimator:
\begin{equation*}
    \sigma_{\bar Y, \bar Y_{a_t'}}=\frac{\sqrt{\left((4 m-8) \zeta_{1}+2 \zeta_{2}\right) /(m-1)}}{\left(m^{2}-m\right)^{-1} \sum_{i \neq j} 4 K-\kappa\left(\hat y_i,\hat y_j\right)-\kappa\left(\hat y_{a_t',i},\hat y_{a_t',j}\right)},
\end{equation*}
where $\zeta_1$ and $\zeta_2$ are standard variance components of the MMD, with more details of the estimator in Appendix.

Therefore, we have the testing threshold for the null hypothesis $H_{0} : \overline{\text{N-TE}}\left(f_Y(\cdot),do(a_t \rightarrow a_t')\right)\leq \epsilon$ with $\epsilon\in(0,1)$ as 
\begin{equation}
\label{eq:test}
    \tau_{\alpha}=\epsilon+\frac{\sigma_{\bar Y, \bar Y_{a_t'}}\mathcal{N}_{1-\alpha}}{\sqrt{m}},
\end{equation}
where $\mathcal N_{1-\alpha}$ is the $(1-\alpha)$-quantile of the standard normal distribution. Based on Eq~\eqref{eq:test}, we define our test function
\begin{equation*}
    t(\hat Y, \hat Y_{a'_t}, \kappa)=\mathbf{1}\left[\overline{\text{N-TE}}\left(f_Y(\cdot),do(a_t\rightarrow a_t')\right)>\tau_{\alpha}\right] ,
\end{equation*}
where $\mathbf 1(\cdot)$ is the indicator function. We reject the null hypothesis if the test statistic exceeds the threshold, indicating there is significant evidence that the two distributions differ. Otherwise, we fail to reject the null hypothesis. Based on Theorem~\ref{thm:asy}, we can directly control the Type I error of the test with a confidence level of $\alpha$.

\begin{Theorem} [N-TE Type-I error testing power]
Under null hypothesis $H_{0}:
{\mathrm{N}\text{-}\mathrm{TE}}\left(f_Y(\cdot),do(a_t\rightarrow a_t')\right)\leq \epsilon$, Type-I error is bounded by $\alpha$, by setting the testing threshold as the $(1-\alpha)$-quantile of the asymptotic null distribution of $\overline{\mathrm{N}\text{-}\mathrm{TE}}\left(f_Y(\cdot),do(a_t=a_t')\right)$ in Theorem~\ref{thm:asy}. That is:
\begin{equation*}
\Pr( \overline{\mathrm{N}\text{-}\mathrm{TE}}\left(f_Y(\cdot),do(a_t\rightarrow a_t')\right)\textgreater \tau_\alpha \mid H_0)\rightarrow \alpha,\quad m\to\infty.
\end{equation*}
\end{Theorem}
Finally, we propose the N-TE testing consistency as follows (The proof will be showed in Appendix~\ref{subsec:A6}).
\begin{Theorem} [N-TE testing consistency]
Under alternative hypothesis $H_{1}:
{\mathrm{N}\text{-}\mathrm{TE}}\left(f_Y(\cdot),do(a_t\rightarrow a_t')\right)\geq \epsilon$,
the test always successfully rejects the null hypothesis based on Theorem~\ref{thm:asy}. That is:
\begin{equation*}
\Pr( \overline{\mathrm{N}\text{-}\mathrm{TE}}\left(f_Y(\cdot),do(a_t\rightarrow a_t')\right)\textgreater \tau_\alpha \mid H_1)\rightarrow 1,\quad m\to\infty.
\end{equation*}
 \end{Theorem}

Therefore, we can write 
\[
\begin{aligned}
    &\Pr( \overline{\text{N-TE}}\left(f_Y(\cdot),do(a_t\rightarrow a_t')\right)\textgreater \tau_\alpha \mid H_1)\\
    =&\Pr(Z>\frac{\tau_{\alpha}-\sqrt{m}\left(\epsilon^{\prime}-\epsilon\right)}{\sigma_{\hat{Y}, \hat{Y}_{a_{t}^{\prime}}}})\to 1, \quad m\to\infty,
\end{aligned}
\]
with standard Gaussian variable $Z\sim\mathcal N(0,1)$.

We now present the counterfactual fairness-closeness testing by taking our $\overline{\text{N-TE}}$ as the measure of distributional closeness.

\begin{algorithm}[H]
  \caption{Counterfactual fairness-closeness testing $\textrm{CF-CLOT}$}
  \label{Algorithm 1}
  \begin{algorithmic}
  \Require Sensitive attributes $A_s$, Observable attributes $A_c$, Bootstrap sample size $m$, A deep model $f_Y(\cdot)$, Testing attribute $a_t$
  \Ensure $H_0$: $f_Y(\cdot)$ is fair with respect to testing attribute $a_t$
  \State \textbf{Step 1:} Finish kernel mean embedding fitting by \eqref{7}.
  \State \textbf{Step 2:} Generate $m$ bootstrap empirical estimator of $\overline{\text{N-TE}}_{\phi}(f_Y(\cdot),do(a_t\rightarrow a_t'))$ by \eqref{26}.
  \State \textbf{Step 3:} Reject $H_0$ with significance level $\alpha$ if the percentage of $\overline{\text{N-TE}}$ satisfies $\overline{\text{N-TE}}_{\phi}(f_Y(\cdot),do(a_t\rightarrow a_t')) > \tau_\alpha$ less than $\alpha$.
  \end{algorithmic}
\end{algorithm}

Based on Definition~\ref{DPOF} and Definition~\ref{icf}, $\textrm{CF-CLOT}$ successfully transforms fairness analysis to DCT based on distribution-based POF, and the testing consistency of NTE ensure unfair sensitive attributes can be successfully tested by CF-CLOT.

\subsection{From the flexibility of $\epsilon$ to the sensitivity of $\textrm{CF-CLOT}$}

Fairness analysis methods must adapt to varying sensitivity requirements due to the inconsistency of fairness metrics across different scenarios. As shown in Definition~\ref{NTE} and Definition~\ref{def:CCFH}, $\epsilon$ represents the maximum acceptable level of unfairness in $\textrm{CF-CLOT}$. Its value can be flexibly and reasonably adjusted to control sensitivity. By tuning $\epsilon$, $\textrm{CF-CLOT}$ can exhibit different sensitivity levels. Based on this, we propose three typical sensitivity levels for $\textrm{CF-CLOT}$.

\subsubsection{Strong sensitivity (Counterfactual
two-sample testing)}
The strong sensitivity of $\textrm{CF-CLOT}$ is required in extremely sensitive scenes, such as racial bias. We define $f_Y$ is fair only when $\mathbb P_n=\mathbb P_{a_t',n}$ in strong sensitivity, which means $\epsilon$ should be set to 0. By Lemma~\ref{lem:srong}, we present our N-TE can also be used to test whether $\mathbb P_n=\mathbb P_{a_t',n}$ (The proof will be showed in Appendix~\ref{subsec:A4}).

\begin{Lemma}
\label{lem:srong}
We have $\mathrm{N}\text{-}\mathrm{TE}(f_Y(\cdot),do(a_t \rightarrow a_t'))=0$ if and only if $\mathbb P_n=\mathbb P_{a_t',n}$ for characteristic kernel $\kappa$.

\end{Lemma}
Hence, the null and alternative hypotheses of the strong sensitivity can be formalized as follows based on Definition~\ref{def:CCFH}:
\begin{equation}
\begin{aligned}
H_{0}:{\mathrm{N}\text{-}\mathrm{TE}}\left(f_Y(\cdot),do(a_t \rightarrow a_t')\right)= 0,\\
H_{1}:{\mathrm{N}\text{-}\mathrm{TE}}\left(f_Y(\cdot),do(a_t \rightarrow a_t')\right) \neq 0.
\end{aligned}
\end{equation}

In this type of task, $\textrm{CF-CLOT}$ can be implemented using MMD, and we will further discuss this special case in Appendix~\ref{A3}.

\subsubsection{Netural sensitivity} 

The netural sensitivity of $\textrm{CF-CLOT}$ means slightly lower sensitivity requirements, such as gender bias, which may more arise from sample size imbalances. To to relax the fairness condition, we can set a relatively lower value for $\epsilon$ for example $0<\epsilon\leq0.1$.

\subsubsection{Weak sensitivity} 

The weak sensitivity of $\textrm{CF-CLOT}$ has more relaxed sensitivity requirements, such as age bias, where the distribution span is typically large. We can set $\epsilon$ to consider unfairness only when there is a significant distributional difference, for example $\epsilon > 0.1$.

\section{Experiments}
In this section, we conduct extensive experiments using real-world datasets to validate the effectiveness of $\textrm{CF-CLOT}$. We design two baseline models for comparison, measuring prediction performance through accuracy and evaluating the unfairness of $do(a_t\rightarrow a_t')$ by examining the difference in prediction performance (The selection of sensitive metrics will be discussed in the Appendix~\ref{SSS}.). We also test $\textrm{CF-CLOT}$ with various values of $\epsilon$ to assess its sensitivity. In addition to conventional fairness datasets, we evaluate $\textrm{CF-CLOT}$ on high-dimensional image data, selecting models with relaxed unfairness to further assess its performance. Finally, we compare $\textrm{CF-CLOT}$ with a state-of-the-art structural causal model(SCM)-based causal fairness analysis model to demonstrate that $\textrm{CF-CLOT}$ achieves state-of-the-art testing performance.

\textbf{Baselines.} As in similar works in recent years and widely recognized evaluation methods \cite{chouldechova2017fair,mehrabi2021survey,martinez2023counterfactual,caton2024fairness}, we propose two baseline models: 1. {\bf SAA}, which trains the model using a set of all attributes to complete the prediction; 2. {\bf SSA}, which uses a set of all attributes except the sensitive attributes. Our model, named {\bf SFA}, excludes all sensitive attributes that are identified as unfair by $\textrm{CF-CLOT}$. To assess the sensitivity of $\textrm{CF-CLOT}$, we set up three versions of {\bf SFA-$\epsilon$}: {\bf SFA-$\epsilon$=0} (strong sensitivity), {\bf SFA-$\epsilon$=0.1} (neutral sensitivity), and {\bf SFA-$\epsilon$=0.3} (weak sensitivity) to conduct an experiment on $\epsilon$ flexibility.

\begin{figure*}
    \centering
    \includegraphics[width=1\linewidth]{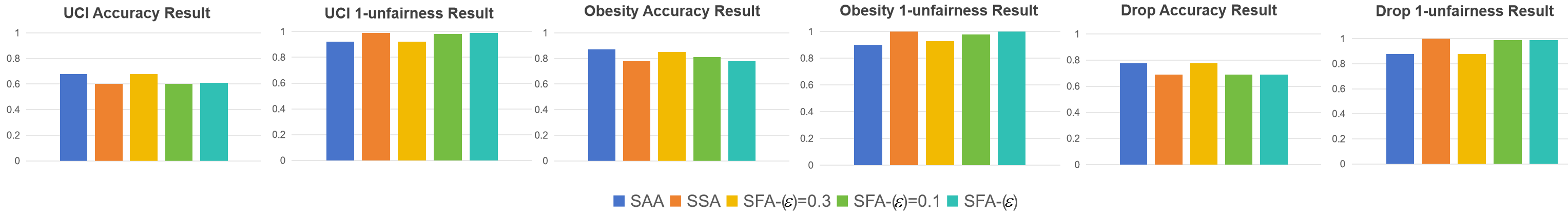}
    \caption{Results of CNN}
    \label{1}
\end{figure*}
\begin{figure*}
    \centering
    \includegraphics[width=1\linewidth]{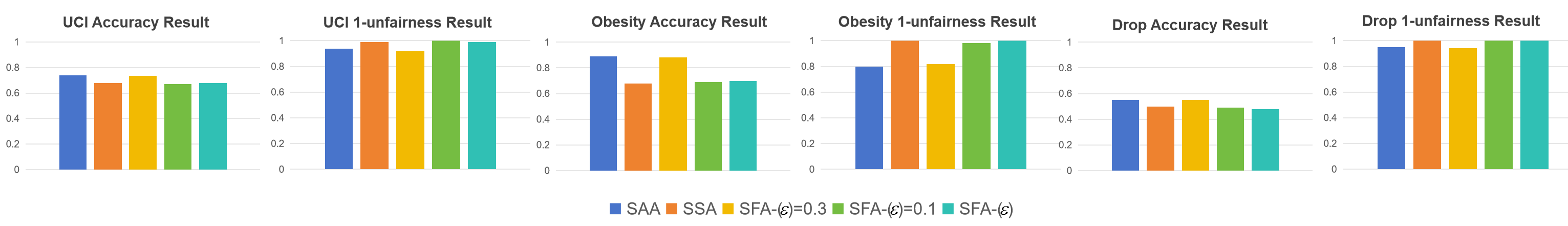}
    \caption{Results of LSTM}
    \label{22}
\end{figure*}
\begin{figure*}
    \centering
    \includegraphics[width=1\linewidth]{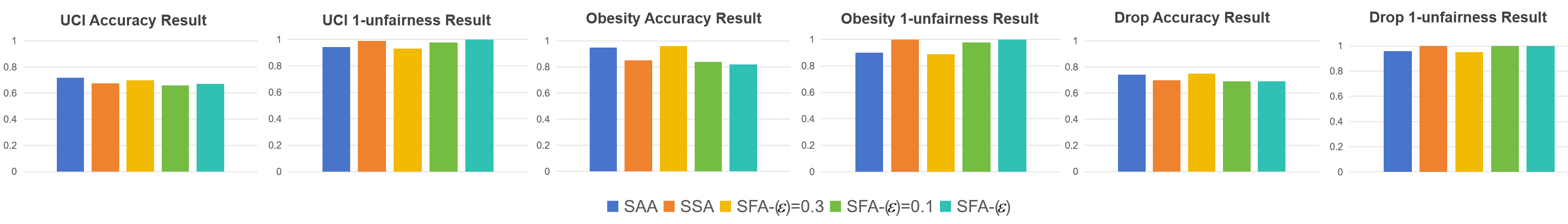}
    \caption{Results of Transformer}
    \label{3}
\end{figure*}
\subsection{Real-world dataset experiment across different models}
In this section, we use $\textrm{CF-CLOT}$ to test the dataset from UCI, which is widely trusted and applied, and can better reflect the discrimination phenomenon that exists in the real world.
\subsubsection{Experiment setting}
\textbf{Datasets and sensitive attributes.} 
We regard sex, parent's job and parent's education levels as the sensitive attributes in the UCI Student Performance Dataset(UCI) \cite{misc_student_performance_320} to predict students' performance in Mathematics.Estimation of Obesity dataset(Obesity) \cite{misc_estimation_of_obesity_levels_based_on_eating_habits_and_physical_condition__544} includes data for the estimation of obesity levels, we regard gender and age as sensitive attributes. We also regard  gender, parents' occupation and parents' qualifications as sensitive attributes in Students' Dropout and Academic Success dataset(Drop) \cite{misc_predict_students'_dropout_and_academic_success_697}.

\textbf{Tested models.} We apply widely used convolutional neural networks(CNN), long short-term memory networks(LSTM), and transformers, as the models to be tested. We use LSTM as a representative of typical recurrent neural network structures , while transformer serves as a representative of self-attention mechanism models.

\subsubsection{Analysis}
As shown in the Figure~\ref{1},~\ref{22} and~\ref{3}, SAA achieved the best accuracy results across all datasets, while SFA-$\epsilon$=0.3 performed similarly, except for the Obesity dataset. In contrast, SSA, SFA-$\epsilon$=0.1, and SFA-$\epsilon$ yielded similar but poorer results. From a fairness perspective, SSA, SFA-$\epsilon$=0.1, and SFA-$\epsilon$ achieved fairness, while SAA and SFA-$\epsilon$=0.3 exhibited comparable degrees of unfairness. Based on the density plot, we can conclude the following:

1. Across different datasets, $\textrm{CF-CLOT}$ successfully identified unfair sensitive attributes, demonstrating it has achieved fairness analysis by DCT with testing consistency. The similar results of SFA-$\epsilon$=0.1, SFA-$\epsilon$, and SSA suggested that the features used for training were the same, demonstrating that $\textrm{CF-CLOT}$ effectively identified the sensitive attributes we set as unfair.

2. By different values of $\epsilon$, $\textrm{CF-CLOT}$ demonstrates its sensitivity in fairness analysis. The different results of SFA-$\epsilon$, SFA-$\epsilon$=0.1, and SFA-$\epsilon$=0.3 highlighted how adjusting $\epsilon$ can fine-tune the sensitivity of $\textrm{CF-CLOT}$. The attributes considered fair in the $\epsilon$=0.3 scenario were deemed unfair in the $\epsilon$=0.1 and $\epsilon$=0 scenarios, illustrating the varying sensitivities. In the Obesity experiment, the results of SFA-$\epsilon$=0.3 and SAA differed, validating that certain sensitive attributes considered unfair in the SFA-$\epsilon$=0.3 scenario were identified as such. It further demonstrated the effectiveness of the weak sensitivity setting.

\subsection{High dimensional dataset experiments}

The excellent performance of $\textrm{CF-CLOT}$ on real-world data demonstrated its effectiveness in testing the fairness of deep models. Moving forward, we will conduct fairness testing experiments in real-world discrimination scenarios, involving high-dimensional data and more complex tasks, to further validate the effectiveness of $\textrm{CF-CLOT}$.

\subsubsection{Facial expression recognition (FER)}

\begin{figure}
    \centering
    \includegraphics[width=1\linewidth]{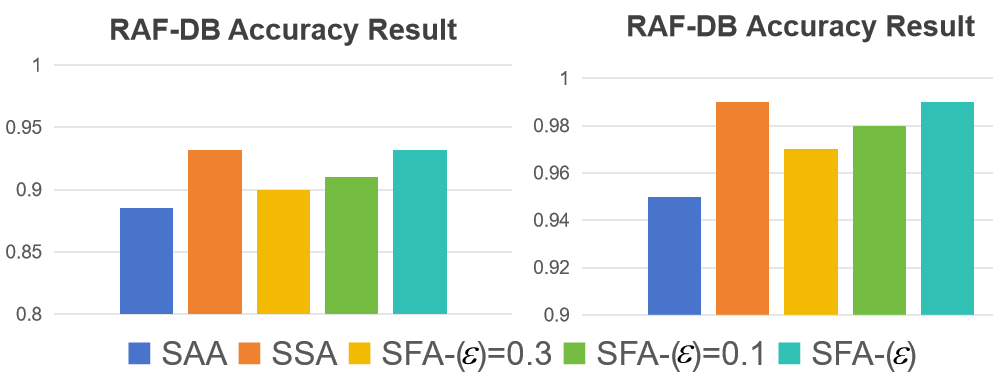}
    \caption{Accuracy and unfairness result for RAF-DB}
    \label{fig:enter-label}
\end{figure}

Facial Emotion Recognition (FER) is widely applied in real-world scenarios. However, discrimination remains prevalent in contemporary FER systems. For example, the prediction accuracy of many commercial FER systems, such as Amazon, Face++, Microsoft, and Sighthound, is significantly lower for elderly individuals compared to other age groups \cite{kim2021age}. Additionally, deep models often exhibit gender bias in their accuracy \cite{domnich2021responsible}. we focus on the issue of emotion discrimination in FER, specifically the tendency to detect positive expressions (such as happiness) more accurately than negative expressions (such as fear).

\textbf{Experiment setting}: We use the Real-world Affective Face Database (RAF-DB) \cite{li2017reliable}, a large-scale facial expression dataset containing 29,672 real-world images labeled with seven emotional categories: Surprise, Happy, Neutral, Disgust, Fear, Sad, and Angry. The RAC-RSL model \cite{zhang2024causaldistillationalleviatingperformance} will be the model under test, owing to its excellent performance in debiasing tasks. In this experiment, we treat the labels of Disgust, Fear, Sad, and Angry as sensitive attributes, and these will be analyzed without the model's awareness of the results.

\textbf{Analysis}:
As shown in Figures 3, unfairness in FER typically manifested as below-average recognition accuracy, meaning that the lower the unfairness, the higher the corresponding accuracy should be. Most of the experimental results aligned with this observation, with SSA and SFA-$\epsilon$ being tested as fair, which further demonstrated that \textbf{$\textrm{CF-CLOT}$ successfully identified unfair sensitive attributes.} Additionally, the performance of SFA-$\epsilon$, SFA-$\epsilon$=0.1, and SFA-$\epsilon$=0.3 varied, where SFA-$\epsilon$ performed the best in terms of fairness and SFA-$\epsilon$=0.3 exhibited the highest degree of unfairness. This discrepancy suggested that the sensitive attributes identified as unfair by the three models were different, likely due to the greater complexity of the images and the RAC-RSL model's strong anti-bias capabilities, which made the N-TE results more distinct compared to previous experiments. In conclusion, \textbf{this further validated the sensitivity of $\textrm{CF-CLOT}$.} Moreover, the excellent performance on image data confirmed that \textbf{$\textrm{CF-CLOT}$ effectively addressed the challenge of conducting causal fairness analysis on high-dimensional data.}

\subsubsection{Credit risk assessment (CRA)}
\begin{figure}
    \centering
    \includegraphics[width=1\linewidth]{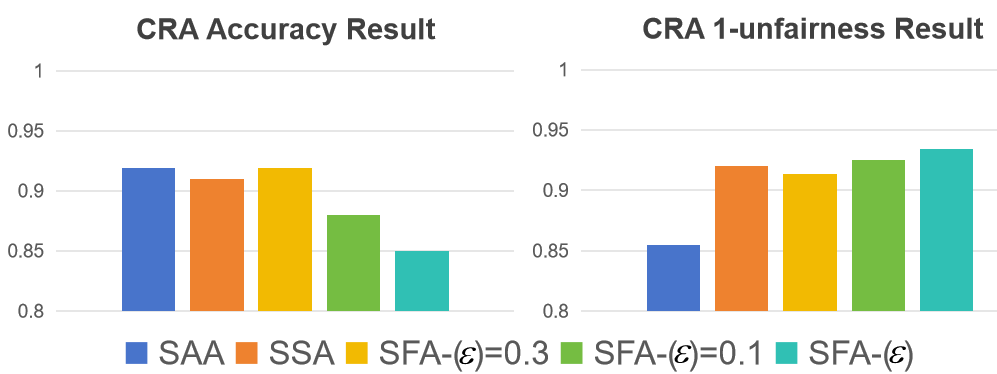}
    \caption{Accuracy and unfairness result for CRA}
    \label{fig:enter-label}
\end{figure}

Statistical bias frequently arises in tasks involving the analysis of demographic attributes, where predictions are unfairly skewed toward specific sensitive attributes such as gender, age, and race \cite{xu2020investigatingbiasfairnessfacial}. One example is credit risk assessment (CRA), which involves predicting the likelihood that a borrower will default on a loan. In such tasks, there is often a bias where relatively older groups are assessed as having a higher risk of default \cite{shi2022machine}.

\textbf{Experiment settings}: We utilize the Credit Risk Dataset, which contains 11 features
related to the repayment capability of 32,581 borrowers. Like some similar studies \cite{zuo2023counterfactualfairnesspartiallyknown,zuo2024interventional}, we treat specific age (\textit{23} (first quantiles) and \textit{30} (third quantiles)) and older age groups (which are \textit{over 40} and \textit{over 50}) as different sensitive attributes. 

\textbf{Analysis}: As shown in Figure 4, the results aligned with the previous experimental conclusions, with SSA and SFA-$\epsilon$ achieving the fairest performances. SFA-$\epsilon$, SFA-$\epsilon$=0.1, and SFA-$\epsilon$=0.3 also demonstrated varying performances, where SFA-$\epsilon$ was the fairest and SFA-$\epsilon$=0.3 exhibited the highest degree of unfairness.

In summary, 

\textbf{1. $\textrm{CF-CLOT}$ successfully identified unfair sensitive attributes on diverse datasets and deep models, demonstrating it has achieved fairness analysis by DCT with testing consistency.}

\textbf{2.$\textrm{CF-CLOT}$ demonstrate the sensitivity in fairness analysis by the fexiblilty of $\epsilon$.} 

\subsection{Comparison with SCM approach}

We compare the $\textrm{CF-CLOT}$ with State-of-art SCM-based fairness testing method \textit{IFair} \cite{zuo2024interventional}. \textit{IFair} involves modeling fair predictions using partial directed acyclic graphs (PDAGs), causal DAGs learned from observational data combined with domain knowledge, reducing the impact of missing key information on the construction of SCMs. \textit{IFair} successfully built SCM for UCI Student Performance Dataset \cite{misc_student_performance_320} and Credit Risk Dataset (details will be showed in Appendix). We set up the same experimental models as \textit{IFair} and use SFA-$\epsilon$'s results as representation of $\textrm{CF-CLOT}$ with \textit{IFair}:
    
\begin{figure}[H]
    \centering
    \includegraphics[width=1\linewidth]{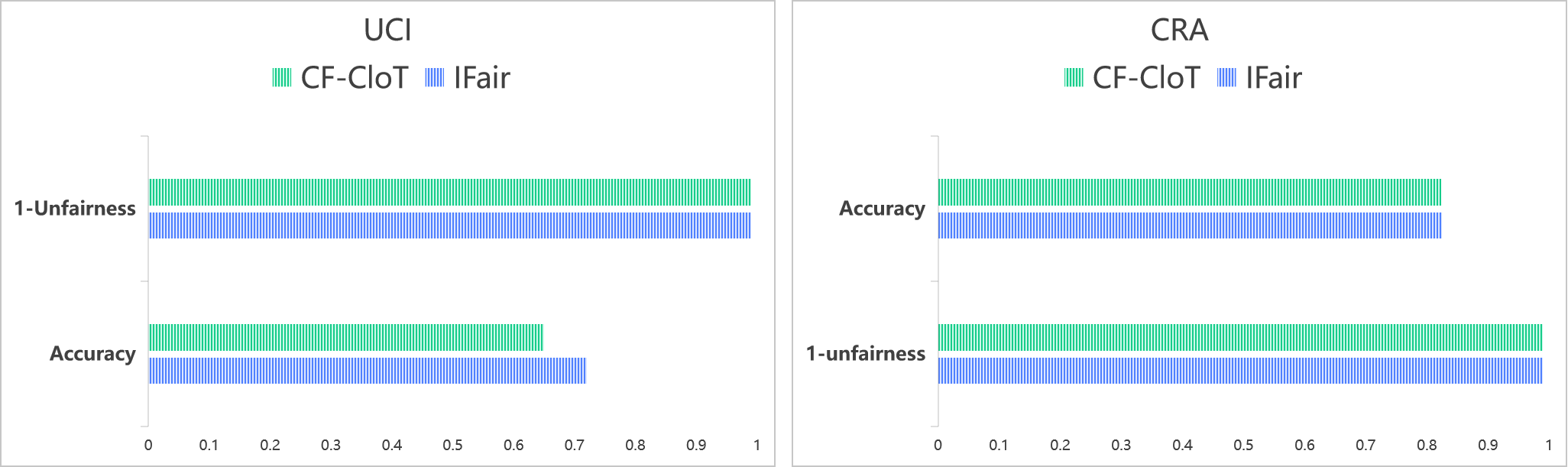}
    \caption{Comparison with $\epsilon$-IFair}
    \label{fig:enter-label}
\end{figure}

As shown in Figure 5, although the effect of $\textrm{CF-CLOT}$ was slightly inferior to that of the SCM method on both datasets, the difference was minimal. This proves that under the same settings, $\textrm{CF-CLOT}$ and $\textit{IFair}$ detected the same unfair sensitive attributes. This further validates the consistency of $\textrm{CF-CLOT}$, equating the sensitivity of distribution testing to fairness analysis.

\section{Conclusion}

In this paper, we introduce distribution-based POF to transform fairness analysis into DCT, and propose our method $\textrm{IoF-CloT}$. We define counterfactual closeness fairness to align closeness with fairness. By introducing N-TE as the statistic for $\textrm{IoF-CloT}$, the testing consistency is established ro ensure a trustworthy result. By apply a flexible and reasonable $\epsilon$ as the fairness confidence, $\textrm{IoF-CloT}$ demonstrate the sensitivity in fairness analysis. $\textrm{CF-CLOT}$ has successfully tested unfair sensitive attributes in extensive experiments across various real-world scenarios, validating the consistency of its testing performance. To summarize, $\textrm{IoF-CloT}$ pioneers the transformation of fairness analysis into DCT, paving the way for further exploration of fairness analysis using distribution-based POF

\bibliographystyle{named}
\bibliography{sample-base}
\newpage

\appendix
\section{Appendix}

\subsection{The explanation of introduction}
\label{into}

\begin{Lemma}
\label{lem:into}
Let $\mathcal Z=\{\boldsymbol z_1,\boldsymbol{z}_2,\ldots,\boldsymbol{z_n}\}\subseteq\mathbb{R}^d$ be a metric space, and let $\mathbb{P}_n$,$\mathbb{Q}_n$ be two Borel probability measures defined on $\mathcal Z$. Then $\mathbb{P}_n = \mathbb{Q}_n$ if and only if $E_z(f(z)=E_{z'}(f(z'))$ for all $f\in C(\mathcal Z)$, where $C(\mathcal Z)$ is the space of bounded continuous functions on $\mathcal Z$ \cite{gretton2012kernel}.

\end{Lemma}

\subsection{Maximum mean discrepancy}
\label{subsec:MMD}
The MMD is a typical kernel-based distance between two distributions \cite{gretton2012kernel}. Let $\mathbb{P}$ and $\mathbb{Q}$ represent two probability measures based on an instance space $\mathcal{X}\subseteq \mathbb{R}^d$. Given $u,v\in\mathcal X$, Let $\kappa:\mathcal{X} \times \mathcal{X} \rightarrow \mathbb{R}$ be the PDS kernel with respect to the corresponding reproducing kernel Hilbert space $(\mathcal{H_{\kappa}},\langle\cdot,\cdot\rangle_{\mathcal{H_{\kappa}}})$, where $\kappa(\cdot,u),\kappa(\cdot,v)\in \mathcal{H_{\kappa}}$ and $\langle \kappa(\cdot,u),\kappa(\cdot,v)\rangle=\kappa(u,v)$. We assume that there exists $K>0$, such that $0	
\leq \kappa(u,v) \leq K$ for any $u,v\in\mathcal X$. The kernel mean embedding of $\mathbb{P}$ and $\mathbb{Q}$ can be defined as:
 \begin{equation}
 \label{2}
 \begin{aligned}
  \mu_\mathbb{P}=E_{u\sim\mathbb{P}}[\kappa(\cdot,u)],\mu_\mathbb{Q}=E_{v\sim\mathbb{Q}}[\kappa(\cdot,v)].
 \end{aligned}
 \end{equation}
 Then the MMD of $\mathbb{Q}$ and $\mathbb{O}$ w.r.t. $\kappa $ is:
  \begin{equation}
 \begin{aligned}
  &\quad \text{MMD}^2(\mathbb{P},\mathbb{Q},\kappa)\\
  &=||\mu_\mathbb{P}-\mu_\mathbb{Q}||_{\mathcal{H_{\kappa}}}^2\\
  &=E_{u\sim\mathbb{P},v\sim\mathbb Q}\langle \kappa(\cdot,u)-\kappa(\cdot,v),\kappa(\cdot,\boldsymbol{u})-\kappa(\cdot,\boldsymbol v)\rangle_{\mathcal{H_{\kappa}}}\\
  &=
  E_{u\sim\mathbb{P},v\sim\mathbb Q}[\kappa(u,\boldsymbol{u})+\kappa(v,\boldsymbol{v})-2\kappa(u,v)]\\
  &\in[0,2K].
 \end{aligned}
 \end{equation}
where $\boldsymbol u$ (resp. $\boldsymbol{v}$) is an i.i.d. copy of $u\sim \mathbb Q$ (resp. $v\sim\mathbb O$).

\subsection{U-statistic}
U-statistic \cite{serfling2009approximation,vershynin2018high} is a type of statistic widely used to estimate parameters, especially in nonparametric statistics. It was proposed by probability theorist and statistician Wassily Hoeffding. U-statistic is often used to construct unbiased estimates of certain parameters, especially when the samples are independent and identically distributed.

\begin{Definition}[U-statistic]
Given ramdom sample $x_1,x_2,...,x_n$ from some distribution, let $h(x_1,x_2,...,x_m)$ represents a symmetric kernel function where $m\leq n$. The U-statistic is defined as:
\begin{equation}
U_n=\frac{1}{(\substack{n \\m})}\sum_{1\leq i_1 \leq i_2...\leq i_n\leq n}h(x_{i_1},x_{i_2},...,x_{i_n}),
\end{equation}
\end{Definition}
where $ (\substack{n \\m}) $ is the combinatorial number, which means all possible combinations of selecting $m$ samples from $n$ samples. In our setting, $ (\substack{n \\m}) $ is the number of ways to choose $m$ distinct indices from $n$.

In practical applications, a typical example is the estimation of the sample mean. The sample mean can be written as a U-statistic:
\begin{equation}
U_n=\frac{1}{n}\sum_{i=1}^n x_i,
\end{equation}
here symmetric kernel function $h(x)=x$ and $m=1$.

Then based on the sub-Gaussian property of bounded functions (please refer to 
\cite{vershynin2018high} for more details about concentration inequalities), we present the large deviation for U-statistic as follows:
\begin{Theorem}[Large deviation for U-statistic]
If the function $h$ is bounded, i.e. $a\leq h(x_1,x_2,...,x_m) \leq b$, we have
\begin{equation}
\mathrm{Pr}(|U_n-\eta|\geq t) \leq 2 \mathrm{exp} \left(-2\lfloor n/m\rfloor t^2/(b-a)^2\right),
\end{equation}
where $\eta=E\left[h(x_{i_1},x_{i_2},...,x_{i_n})\right]$.
\end{Theorem}

\subsection{Special case: Counterfactual two-sample testing } \label{A3}
Due to the characteristics of strong sensitivity, the model is considered fair iff $\mathbb P_n=\mathbb P_{a_t',n}$. Therefore, $\textrm{CF-CLOT}$ is equivalent to two sample testing in this setting. Now we give the definition of the corresponding MMD version as a special case.

\begin{Definition}[MMD treatment effect (M-TE)] 
Given the tested deep model $f_Y(\cdot)$ and sensitive attribute $a_t\in A_s$, suppose $\hat{Y}$ represents factual potential outcomes distribution and $\hat{Y}_{a_t'}$ represents the counterfactual one. The $\mathrm{M}$-$\mathrm{TE}$ of $do(a_t\rightarrow a_t')$ is
\begin{equation}
\begin{aligned}
&\textrm{M-TE}(f_Y(\cdot),do(a_t\rightarrow a_t'))=||\mu_{\hat{Y}}-\mu_{\hat{Y}_{a_t'}}||_{\mathcal{H_{\kappa}}}^2\\
&=E_{\hat Y,\hat Y_{a_t'}}[\kappa(\hat{y},\boldsymbol{\hat{y}})+\kappa(\hat{y}_{a_t'},\boldsymbol{\hat{y}}_{a_t'})-2\kappa(\hat{y},\hat{y}_{a_t'})],
\end{aligned}
\end{equation}
\end{Definition}
and it is also clear that $\text{M-TE} \in [0, 2K]$. Here, the value of M-TE approaches 0 when the $f_Y(\cdot)$ is not biased toward $a_t$ .

We further introduce the empirical estimator of M-TE as :
\begin{equation}
\label{13}
\begin{aligned}
&\quad\overline{\text{M-TE}}(f_Y(\cdot),do(a_t\rightarrow a_t'))=\\
&\quad\sum_{i\neq j}\kappa(\hat{y}_i,\hat{y}_j)+\kappa({\hat{y}}_{a_t',i},{\hat{y}}_{a_t',j})-\kappa(\hat{y}_i,\hat{y}_{a_t',j})-\kappa(\hat{y}_{a_t',i},\hat{y}_j).
\end{aligned}
\end{equation}

together with its bootstrap version:

\begin{equation}
\begin{aligned}
&\quad\overline{\text{M-TE}}_\phi(f_Y(\cdot),do(a_t=a_t'))\\
=&\quad\sum_{i\neq j} \phi_{ij} \left(\kappa(\hat{y}_i,\hat{y}_j)+\kappa({\hat{y}}_{a_t',i},{\hat{y}}_{a_t',j})-\kappa(\hat{y}_i,\hat{y}_{a_t',j})-\kappa(\hat{y}_{a_t',i},\hat{y}_j)\right),
\end{aligned}
\end{equation}

\subsection{Detailed Proof of Lemma~\ref{lem:srong}}
\label{subsec:A4}
We begin with a well-known theorem as follows.
\begin{Theorem}
\label{thm:wellknown}
\cite{gretton2012kernel} Denote by $\mathbb{P}$ and $\mathbb{Q}$  two Borel probability measures over space $X\subseteq\mathbb{R}^d$. Let $\kappa:X \times X \rightarrow \mathbb{R}$. be a characteristic kernel. Then $\mathrm{MMD}^2(\mathbb{P},\mathbb{Q},\kappa)=0$ if and only if $\mathbb{P}=\mathbb{Q}$.
\end{Theorem}

We now present the proof of Lemma~\ref{lem:srong} as follows, by taking $\mathbb P=\mathbb P_n$ and $\mathbb Q=\mathbb P_{a_t',n}$ in Theorem~\ref{thm:wellknown}.

\begin{proof}
Suppose that random variables $\hat Y\sim\mathbb P_n$ and $\hat Y_{a'_t}\sim\mathbb P_{a_t',n}$, then we have
\begin{equation}
\begin{aligned}
&\text{N-TE}(f_Y(\cdot),do(a_t\rightarrow a_t'))\\
=&\frac{E_{\hat Y,\hat Y_{a_t'}}[\kappa(\hat{y},\boldsymbol{\hat{y}})+\kappa(\hat{y}_{a_t'},\boldsymbol{\hat{y}}_{a_t'})-2\kappa(\hat{y},\hat{y}_{a_t'})]}{4K-E_{\hat Y,\hat Y_{a_t'}}[\kappa(\hat{y},\boldsymbol{\hat{y}})+\kappa(\hat{y}_{a_t'},\boldsymbol{\hat{y}}_{a_t'})]}\\
=&\frac{||\mu_{\hat{Y}}-\mu_{Y_{a_t'}}||^2_{\mathcal H_\kappa}}{4K-||\mu_{\hat{Y}}||^2_{\mathcal H_\kappa}-||\mu_{\hat{Y_{a_t'}}}||^2_{\mathcal H_\kappa}}\\
=&\frac{\text{M-TE}(f_Y(\cdot),do(a_t\to a_t'))}{4K-||\mu_{\hat{Y}}||^2_{\mathcal H_\kappa}-||\mu_{\hat{Y_{a_t'}}}||^2_{\mathcal H_\kappa}}\\
=&\frac{\text{MMD}^2(\mathbb P_n,\mathbb P_{a_t',n},\kappa)}{4K-||\mu_{\hat{Y}}||^2_{\mathcal H_\kappa}-||\mu_{\hat{Y_{a_t'}}}||^2_{\mathcal H_\kappa}}
\end{aligned}
\end{equation}
It is evident that $4K-||\mu_{\hat{Y}}||^2_{H_\kappa}-||\mu_{\hat{Y_{a_t'}}}||^2_{H_\kappa}>0$. Consequently, $ \text{N-TE}(f_Y(\cdot),do(a_t\rightarrow a_t'))=0$ if and only
if $\mathbb P_n=\mathbb P_{a_t',n}$ for characteristic kernels. This completes the proof.    
\end{proof}

\subsection{Detailed Proof of Themorem 1}
\label{subsec:A5}
We begin with the empirical estimator of M-TE as.
\begin{equation}
\label{13}
\begin{aligned}
&\overline{\text{M-TE}}(f_Y(\cdot),do(a_t\rightarrow a_t'))\\
=&\overline{\text{MMD}}^2(\bar Y,\bar Y_{a_t'},\kappa)\\
=&\frac{1}{m(m-1)}\sum_{i\neq j}\kappa(\hat{y}_i,\hat{y}_j)+\kappa({\hat{y}}_{a_t',i},{\hat{y}}_{a_t',j})-\kappa(\hat{y}_i,\hat{y}_{a_t',j})-\kappa(\hat{y}_{a_t',i},\hat{y}_j).
\end{aligned}
\end{equation}

Given this, we introduce a well-known theorem as follows.
\begin{Theorem}
\label{thm:mmd}
Under null hypothesis $H_0'$: $\mathbb P=\mathbb Q$, let $Z_i \sim \mathcal N(0,2)$ and we have:
\begin{center}
$m\overline{\mathrm{MMD}}^2(\bar{Y},\bar{Y}_{a_t'},\kappa)\overset{d}{\rightarrow} \sum_i\lambda_i(Z_i^2-2)$;
\end{center}
here $\lambda_i$ are the eigenvalues of the $\mathbb P$-covariance operator of the centered kernel [\cite{gretton2012kernel}, Theorem 12]. On the other hand, under the alternative $H_1':\mathbb P\neq\mathbb Q$ ,
a standard central limit theorem holds [\cite{serfling2009approximation}, Section 5.5.1]
\begin{center}
$\big( \overline{\mathrm{MMD}}^2(\bar{Y},\bar{Y}_{a_t'},\kappa)- \mathrm{M}\text{-}\mathrm{TE}(f_Y(\cdot),do(a_t\rightarrow a_t') \big) \overset{d}{\rightarrow} \mathcal N(0,\sigma_M^2)$,

$\sigma_M^2:=4E[H_{1,2}H_{1,3}]-4(E[H_{1,2}])^2$,
\end{center}
where $H_{i, j}=\kappa(\hat{y}_i,\hat{y}_j)+\kappa({\hat{y}}_{a_t',i},{\hat{y}}_{a_t',j})-\kappa(\hat{y}_i,\hat{y}_{a_t',j})-\kappa(\hat{y}_{a_t',i},\hat{y}_j)$ and the expectation are taken with
respect to $\hat{y_1},\hat{y_2},\hat{y_3}\sim \mathbb P^3$ and $\hat{y_{a_t',1}},\hat{y_{a_t',2}},\hat{y_{a_t',3}}\sim \mathbb Q^3$.
\end{Theorem}
We now present the proofs of Theorem~\ref{thm:asy} as follows.
\begin{proof}
    Recall the empirical estimator of our N-TE:
\begin{equation}
\begin{aligned}
&m\overline{\text{N-TE}}(f_Y(\cdot),do(a_t \rightarrow a_t'))\\
=&\frac{m\overline{\text{M-TE}}^2(f_Y(\cdot),do(a_t\rightarrow a_t'))}{1/(m^2-m)\sum_{i\neq j}[4K-\kappa(\hat{y}_i,\hat{y}_j) - \kappa({\hat{y}}_{a_t',i},\hat{y}_{a_t',j})]}
\\
=&\frac{m\overline{\text{MMD}}^2(\bar{Y},\bar{Y}_{a_t'},\kappa)}{1/(m^2-m)\sum_{i\neq j}[4K-\kappa(\hat{y}_i,\hat{y}_j) - \kappa({\hat{y}}_{a_t',i},\hat{y}_{a_t',j})]}
    \end{aligned}
\end{equation}
As a U-statistic, it is easy to see that
\begin{equation}
\begin{aligned}
& 1/(m^2-m)\sum_{i\neq j}[4K-\kappa(\hat{y}_i,\hat{y}_j) - \kappa({\hat{y}}_{a_t',i},\hat{y}_{a_t',j})]
\\
& \overset{d}{\rightarrow} 4K -||\mu_{\hat{Y}}||_{\mathcal H_\kappa}^2-||\mu_{\hat{Y}_{a_t'}}||_{\mathcal H_\kappa}^2,
  \end{aligned}
\end{equation}
where $\overset{d}{\rightarrow}$ denotes convergence in probability.

If $\overline{\text{N-TE}}(f_Y(\cdot),do(a_t \rightarrow a_t'))=0$, we have $\mathbb P_n=\mathbb P_{a_t',n}$ from lemma~\ref{lem:srong}, and
\[
m\overline{\text{MMD}}^2(\bar{Y},\bar{Y}_{a_t'},\kappa)\overset{d}{\rightarrow} \sum_i\lambda_i(Z_i^2-2)
\]
from Theorem~\ref{thm:mmd}. Then, by slutsky’s theorem \cite{edition2002probability}, we have
\begin{equation}
\begin{aligned}
m\overline{\text{N-TE}}(f_Y(\cdot),do(a_t \rightarrow a_t')) & \overset{d}{\rightarrow}\frac{\sum_i\lambda_i(Z_i^2-2)}{4K -||\mu_{\hat{Y}}||_{\mathcal H_\kappa}^2-||\mu_{\hat{Y}_{a_t'}}||_{\mathcal H_\kappa}^2}\\
&\overset{d}{\rightarrow}\frac{\sum_i\lambda_i(Z_i^2-2)}{4K -||(\mu_{\hat{Y}}+\mu_{\hat{Y}_{a_t'}})/\sqrt{2}||_{\mathcal H_\kappa}^2},
\end{aligned}
\end{equation}

where $\mu_{\hat{Y}}=\mu_{\hat{Y}_{a_t'}}=(\mu_{\hat{Y}}+\mu_{\hat{Y}_{a_t'}})/2$.

If $\text{N-TE}(f_Y(\cdot),do(a_t \rightarrow a_t'))=\epsilon$ with$ \epsilon \in (0, 1)$, we present the asymptotic distribution of the empirical
estimator in a similar manner, which can be formalized as
\begin{equation}
\begin{aligned}
&\sqrt{m}(\overline{\text{N-TE}}(f_Y(\cdot),do(a_t \rightarrow a_t'))-\epsilon)\overset{d}{\rightarrow}\\
&N\left(0,\frac{4E[H_{1,2}H_{1,3}]-4(E[H_{1,2}])^2}{(4K -||\mu_{\hat{Y}}||_{H_\kappa}^2-||\mu_{\hat{Y}_{a_t'}}||_{H_\kappa}^2)^2}\right)
\end{aligned}
\end{equation}
\end{proof}

\subsection{Detailed Proof of Themorem 3}
\label{subsec:A6}
\begin{proof}
    Conditioned on alternative hypothesis $H_{1}$, we suppose that ${\text{N-TE}}\left(f_Y(\cdot),do(a_t\rightarrow a_t')\right)= \epsilon'$ with $\epsilon'> \epsilon$.

Based on Theorem~\ref{thm:asy}, we have 
\[
\begin{aligned}
    &\sqrt{m}(\overline{\text{N-TE}}\left(f_Y(\cdot),do(a_t\rightarrow a_t')\right)-\epsilon)\mid   \\
   & \text{H}_1\xrightarrow{d}\mathcal N\left(\sqrt{m}(\epsilon'-\epsilon), \sigma_{\hat Y,\hat Y_{a_t'}}^2\right),
\end{aligned}
\]
which completes our proof.
\end{proof}

\subsection{Sensitive attribute setting}
\label{SSS}
The selection of sensitive attributes requires careful consideration. Drawing from previous research in fair machine learning \cite{caton2024fairness,mehrabi2021survey}, commonly recognized sensitive attributes, such as gender and race, should be prioritized \cite{zuo2022counterfactual,zuo2023counterfactualfairnesspartiallyknown,zuo2024interventional}. Additionally, private information \cite{yu2020fairness,ezzeldin2023fairfed,chen2023privacy}, such as family occupation and family history, should also be considered. Given the growing emphasis on deep learning value alignment \cite{wang2024essence,shen2023large}, attributes linked to human social values should also be taken into account.

It is important to note that the selection of sensitive attributes should be guided by the question, "Should this attribute be considered?" rather than "Is this attribute relevant in reality?" \cite{mehrabi2021survey}. For instance, when predicting "grades," while there may be a correlation between "grades" and "gender" in reality, from a fairness perspective, gender should not influence the prediction of student performance. Therefore, in this context, "gender" should be treated as a sensitive attribute.

%\subsection{\textit{IFair}'s causal graph \cite{zuo2024interventional}}
%\label{subsec:A7}
%\begin{figure}[H]
%    \centering
%    \includegraphics[width=1\linewidth]{1736354088917.png}
%    \caption{Causal graphs for the UCI student dataset \cite{zuo2024interventional}}
%    \label{fig:enter-label}
%\end{figure}

%\begin{figure}[H]
%    \centering
%    \includegraphics[width=1\linewidth]{1736354172466.png}
%    \caption{Causal graphs for the Credit risk dataset \cite{zuo2024interventional}}
%    \label{fig:enter-label}
%\end{figure}

\end{document}